\documentclass{article}
\usepackage{spconf,amsmath}

\usepackage{array}
\def\thline{\noalign{\hrule height 1pt}}

\usepackage{bm}

\usepackage[pdftex]{graphicx}

\newcommand{\figref}[1]{Figure \ref{#1}}
\newcommand{\tabref}[1]{Table \ref{#1}}

\newcommand{\secref}[1]{Section \ref{#1}}

\usepackage{subfigure}

\title{Language Modeling with Highway LSTM}
\name{\thanks{$^1$IBM Research - Tokyo, Japan}Gakuto Kurata$^1$, Bhuvana Ramabhadran$^2$, George Saon$^2$, Abhinav Sethy$^2$\thanks{$^2$IBM T.J. Watson Research Center, USA}}
\address{IBM Research AI\\{\small \tt gakuto@jp.ibm.com, \{bhuvana,gsaon,asethy\}@us.ibm.com} }

\begin{document}
%
\maketitle
\begin{abstract}
 Language models (LMs) based on Long Short Term Memory (LSTM) have shown good gains in many
 automatic speech recognition tasks.
 In this paper, we extend an LSTM by adding highway networks inside an LSTM and use the resulting Highway LSTM (HW-LSTM) model for language modeling.
 The added highway networks increase the depth in the time dimension.
 Since a typical LSTM has two internal states, a memory cell and a hidden
 state, we compare various types of HW-LSTM by adding highway
 networks onto the memory cell and/or the hidden state. Experimental
 results on English broadcast news and conversational telephone speech
 recognition show that the proposed HW-LSTM LM improves speech
 recognition accuracy on top of a strong LSTM LM baseline.
 We report 5.1\% and 9.9\% on the Switchboard and CallHome subsets of the Hub5 2000 evaluation, which reaches the best performance numbers reported on these tasks to date.

 
\end{abstract}
\section{Introduction}
\label{sec:intro}

Deep learning based approaches, especially recurrent neural networks
and their variants, have been one of the hottest topics in language
modeling research for the past few years.  Long Short Term Memory
(LSTM) based recurrent language models (LMs) have shown significant
perplexity gains on well established benchmarks such as the Penn Tree
Bank~\cite{zaremba2014recurrent} and the more recent One Billion
corpus~\cite{jozefowicz2016exploring}.  These results validate the
potential of deep learning and recurrent models as being key to further
progress in the field of language modeling.  Since LMs are one of the
core components of natural language processing (NLP) tasks such
as automatic speech recognition (ASR) and Machine Translation (MT),
improved language modeling techniques have sometimes translated to improvements
in overall system performance for these tasks~\cite{xiong2016achieving,mikolov2010recurrent,auli2013joint}.

Enhancements of recurrent neural networks such as deep transition
networks, recurrent highway networks, and fast-slow recurrent neural
networks that add non-linear transformations in the time dimension have
shown superior performance especially in NLP
tasks~\cite{pascanu2013construct,zilly2016recurrent,mujika2017fast}.
Inspired by these ideas, we extend LSTMs by adding highway networks
inside LSTMs and we call the resulting model {\it Highway LSTM
  (HW-LSTM)}.
 The added highway networks further strengthen the LSTM
capability of handling long-range dependencies.  To the best of our
knowledge, this is the first research that uses {\it HW-LSTM} for
language modeling in the context of a state-of-the-art speech recognition task.

In this paper, we present extensive empirical results showing the
advantage of {\it HW-LSTM} LMs on state-of-the-art broadcast news and
conversational ASR systems built on publicly available data.  We
compare multiple variants of {\it HW-LSTM} LMs and analyze the
configuration that achieves better perplexity and speech recognition
accuracy.
We also present a training procedure of a {\it HW-LSTM} LM
initialized with a regular LSTM LM.  Our results also show that the
regular LSTM LM and the proposed {\it HW-LSTM} LMs are complementary
and can be combined to obtain further gains.
The proposed methods were instrumental in reaching the current best reported accuracy on the widely-cited Switchboard (SWB)~\cite{xiong17:_micros_conver_speec_recog_system} and CallHome (CH)~\cite{george17:_englis_conver_telep_speec_recog_human_machin,kurata17:_empir_explor_novel_archit_objec_languag_model} subsets of the NIST Hub5 2000 evaluation testset.

Our paper has three main contributions:
\begin{itemize}
 \item a novel language modeling technique with {\it HW-LSTM}, 
 \item a training procedure of {\it HW-LSTM} LMs initialized with regular LSTM LMs, and
 \item the impact of the above proposed methods in state-of-the-art ASR tasks with publicly available broadcast news and conversational telephone speech data.
\end{itemize}

This paper is organized as follows. We summarize related work in
\secref{sec:related-work} and detail our proposed language modeling
with {\it HW-LSTM} in \secref{sec:highway-lstm}. Next, we confirm the
advantage of {\it HW-LSTM} LMs through a wide range of speech
recognition experiments in \secref{sec:experiments}. Finally, we
conclude this paper in \secref{sec:conclusion}.


\section{Related Work}
\label{sec:related-work}
In this section, we summarize related work that serves as a basis for our proposed method which is described in \secref{sec:highway-lstm}.

\subsection{Highway Networks}
\label{sec:highway}
Highway networks make it easy to train very deep neural networks~\cite{srivastava2015highway}.
The input $x$ is transformed to output $y$ by a highway network with information flows being controlled by transformation gate $g_{T}$ and carry gate $g_{C}$ as follows:
\begin{eqnarray}
 g_{T} &=& \mbox{sigm}(W_{T}x+b_{T}) \nonumber \\
 g_{C} &=& \mbox{sigm}(W_{C}x+b_{C}) \nonumber \\
 y &=& x \odot g_{C} + \mbox{tanh}(Wx+b) \odot g_{T} \nonumber
\end{eqnarray}
$W_{T}$ and $b_{T}$ are the weight matrix and bias vector for the transform gate.
$W_{C}$ and $b_{C}$ are the weight matrix and bias vector for the carry gate.
$W$ and $b$ are the weight matrix and bias vector and non-linearity other than $\tanh$ can be used here.
Highway networks have been showing strong performance in various applications including language modeling~\cite{kim2015character}, image classification~\cite{srivastava2015training}, to name a few.

\subsection{Recurrent Highway Networks}
A typical Recurrent Neural Network (RNN) has one non-linear transformation from a hidden state $h_{t-1}$ to the next hidden state $h_{t}$ given by:
\begin{eqnarray}
 h_{t} &=& \tanh(W_{x}x_{t}+W_{h}h_{t-1}+b) \nonumber
\end{eqnarray}
$x_{t}$ is the input to the RNN at time step $t$, $W_{*}$ are the weight matrices, and $b$ is the bias vector. A
Recurrent Highway Network (RHN) was recently proposed by combining RNN and highway network~\cite{zilly2016recurrent}.
An RHN applies multiple layers of highway networks when transforming $h_{t-1}$ to $h_{t}$.
Multiple layers of highway networks serve as a ``memory''.

\subsection{LSTM}
\label{sec:lstm}
\begin{figure}[t]
  \begin{center}
   \includegraphics[width=0.8\columnwidth]{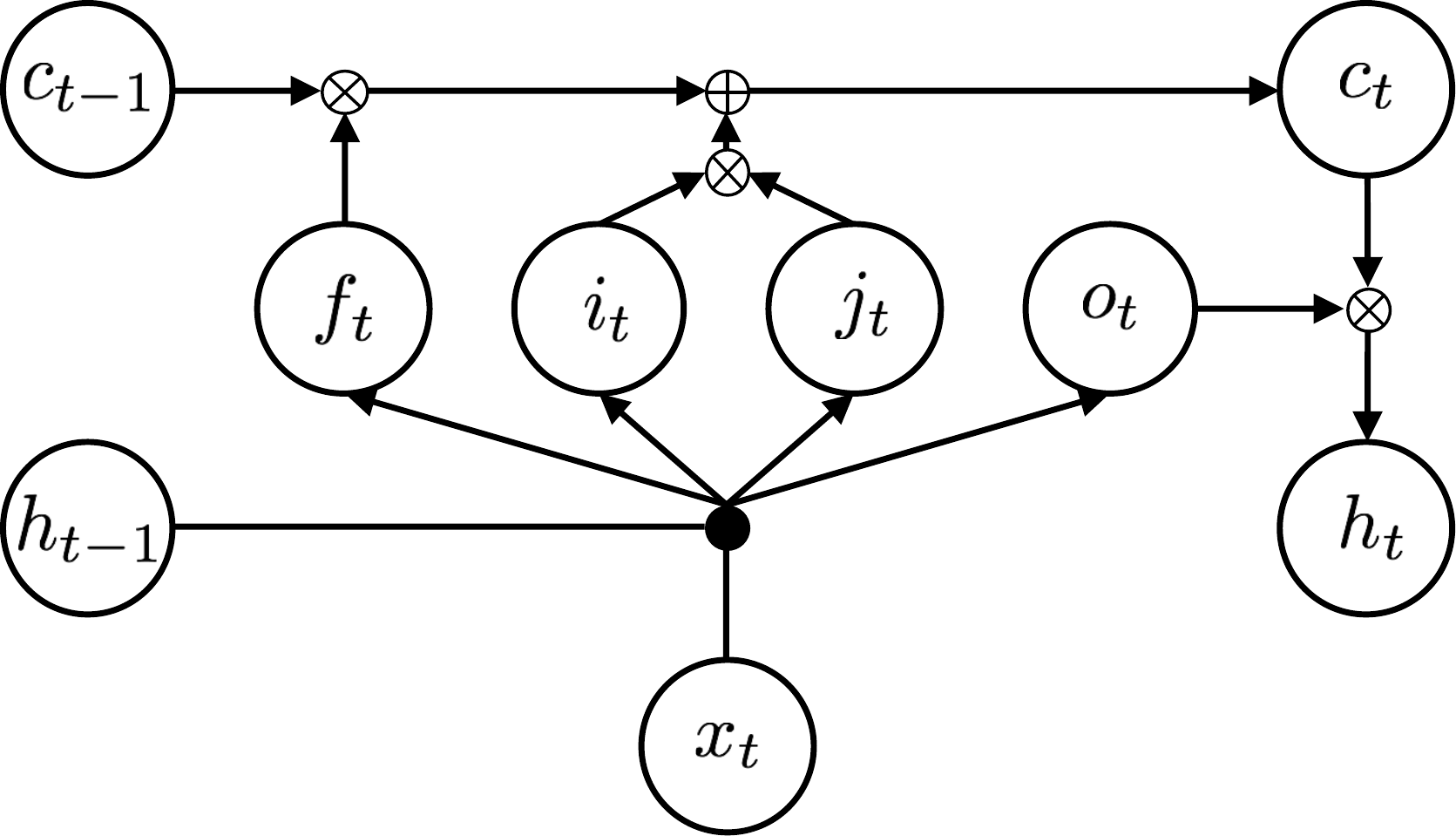}
    \caption{Regular LSTM architecture~\cite{jozefowicz2015empirical}.}
  \label{fig:lstm}
  \end{center}
\end{figure}

An LSTM is a specific architecture of an RNN which avoids the vanishing (or exploding) gradient problem and is easier to train thanks to its internal memory cells and gates.
After exploring a few variants of LSTM architectures,  we settled on the architecture specified below, which is similar to the architectures of \cite{graves2013generating,jozefowicz2015empirical} illustrated in \figref{fig:lstm}.
\begin{eqnarray}
 i_{t} &=& \tanh(W_{xi}x_{t}+W_{hi}h_{t-1}+b_{i}) \nonumber \\
 j_{t} &=& \mbox{sigm}(W_{xj}x_{t}+W_{hj}h_{t-1}+b_{j}) \nonumber \\
 f_{t} &=& \mbox{sigm}(W_{xf}x_{t}+W_{hf}h_{t-1}+b_{f}) \nonumber \\
 o_{t} &=& \mbox{sigm}(W_{xo}x_{t}+W_{ho}h_{t-1}+b_{o}) \nonumber \\
 c_{t} &=& c_{t-1} \odot f_{t} + i_{t} \odot j_{t} \nonumber \\
 h_{t} &=& \tanh(c_{t}) \odot o_{t} \nonumber
\end{eqnarray}
$x_{t}$ is the input to the LSTM at time step $t$, $W_{*}$ are the weight matrices, and $b_{*}$ are the bias vectors.
$\odot$ denotes an element-wise product. $c_{t}$ and $h_{t}$ represent the memory cell vector and the hidden vector at time step $t$.
Note that this LSTM does not have peephole connections.

\section{Highway LSTM}
\label{sec:highway-lstm}

\begin{figure}[t]
  \begin{center}
   \includegraphics[width=\columnwidth]{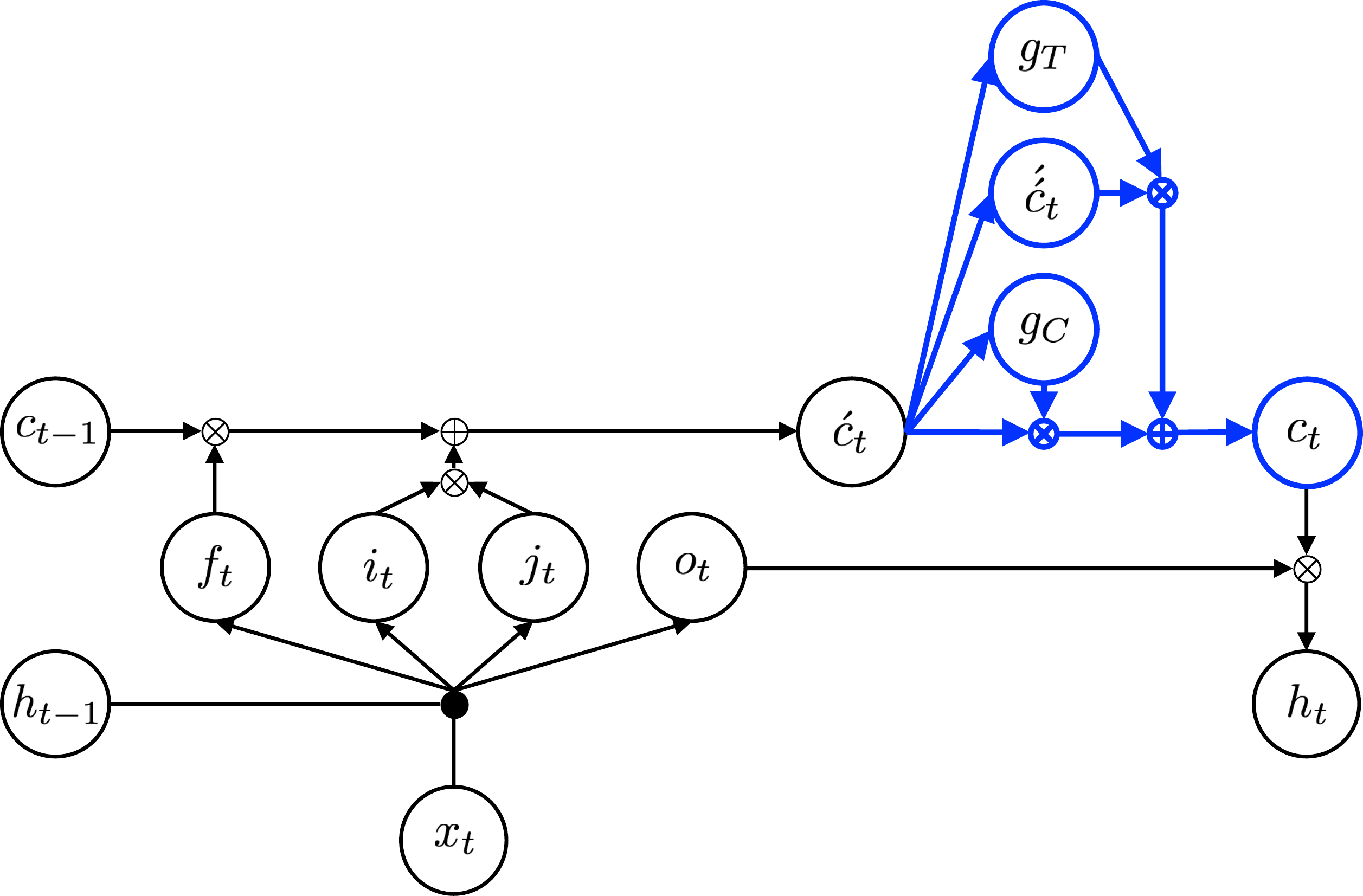}
   \caption{{\bf HW-LSTM-C}: One layer highway network transforms regular LSTM cell $\acute{c}_{t}$ to $c_{t}$ and hidden state $h_{t}$ is calculated on top of transformed memory cell $c_{t}$.}
  \label{fig:hlstm_c}
  \end{center}
\end{figure}

\begin{figure}[t]
  \begin{center}
   \includegraphics[width=\columnwidth]{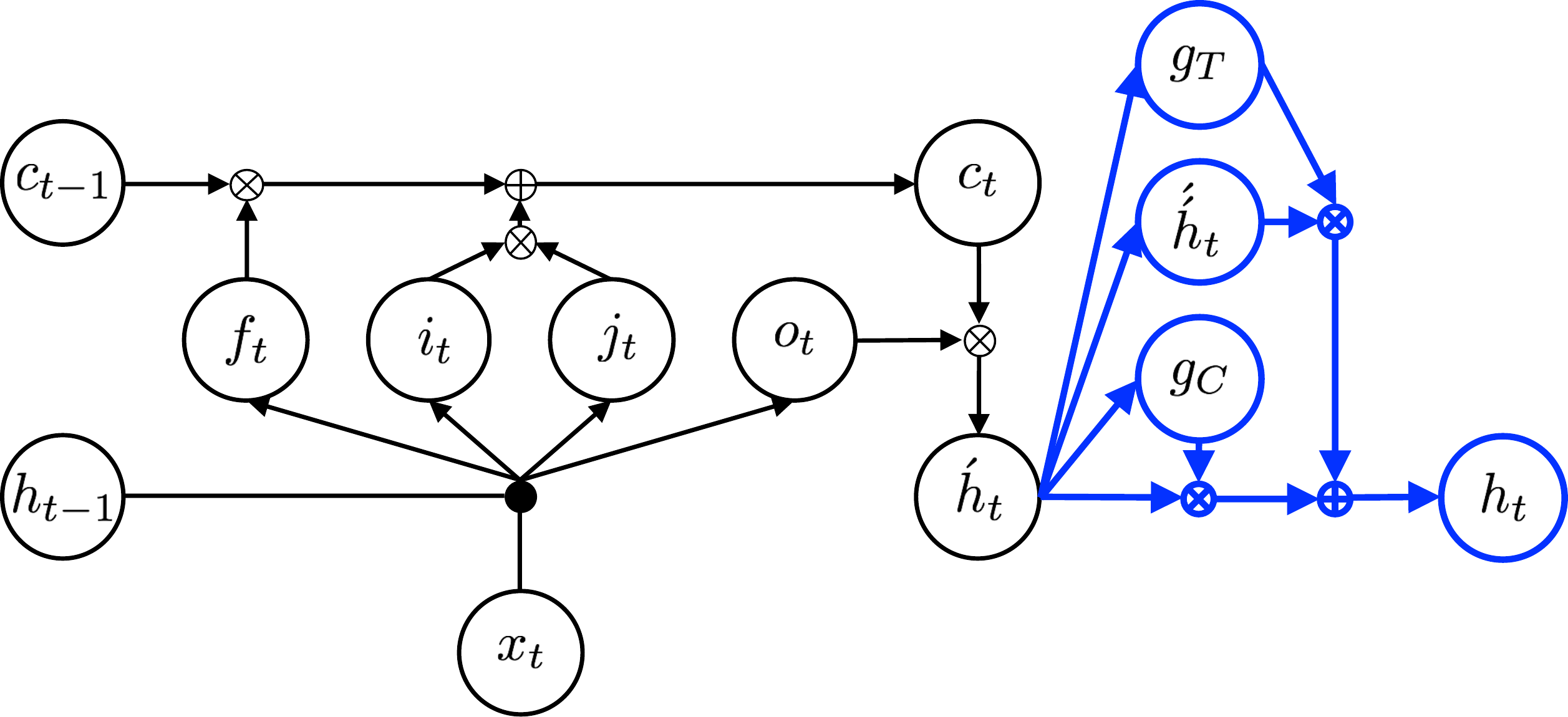}
   \caption{{\bf HW-LSTM-H}: One layer highway network transforms the hidden state of regular LSTM $\acute{h}_{t}$ to $h_{t}$.}
  \label{fig:hlstm_h}
  \end{center}
\end{figure}

\begin{figure}[t]
  \begin{center}
   \includegraphics[width=\columnwidth]{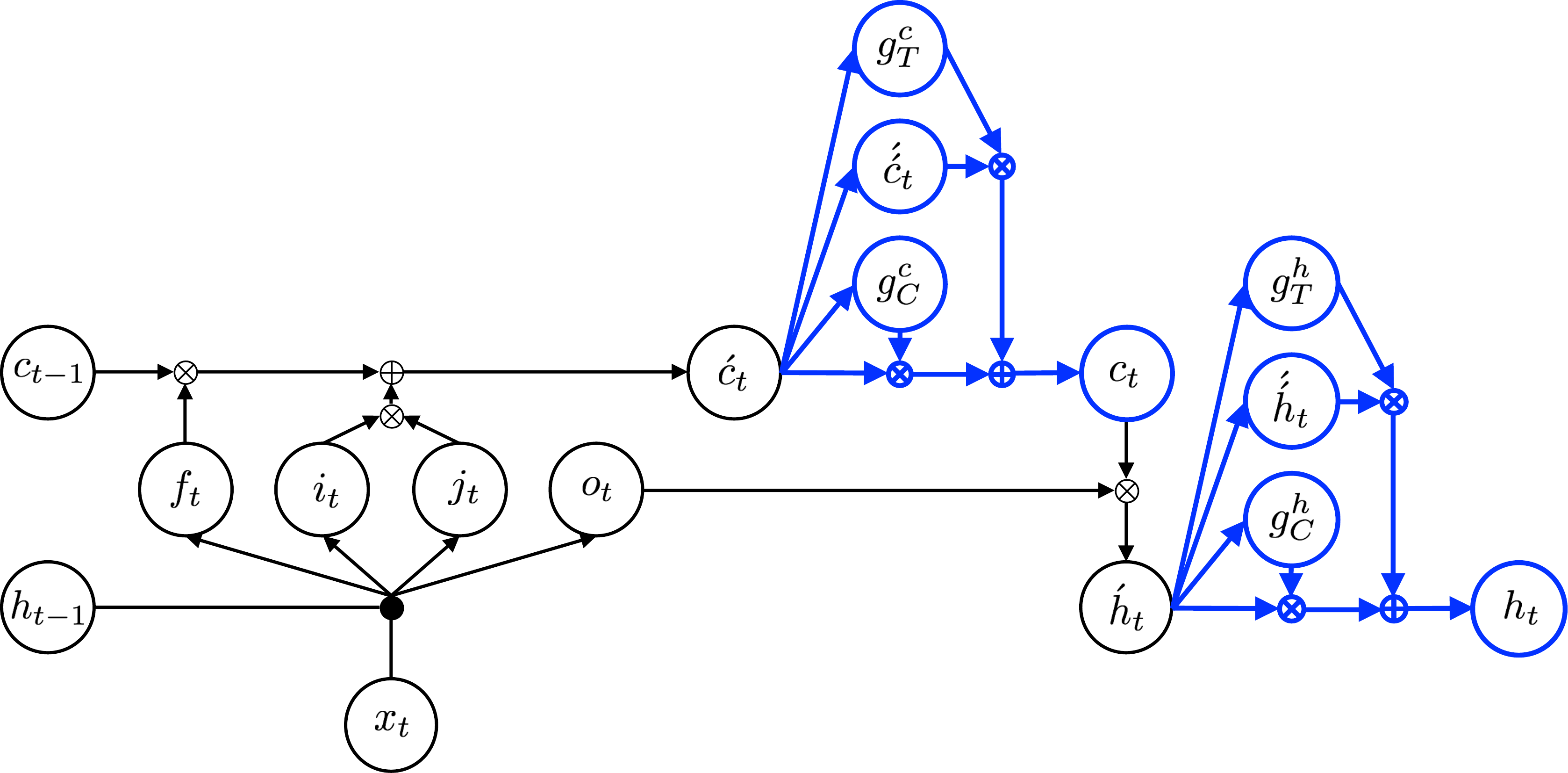}
      \caption{{\bf HW-LSTM-CH}: One layer highway network transforms regular LSTM cell $\acute{c}_{t}$ to $c_{t}$ and hidden state $\acute{h}_{t}$ is calculated on top of transformed memory cell $c_{t}$. Another highway network transforms $\acute{h}_{t}$ to $h_{t}$.}
  \label{fig:hlstm_ch}
  \end{center}
\end{figure}

Inspired by the improved performance achieved by adding multiple layers of highway networks to regular RNNs, we propose {\it Highway LSTM (HW-LSTM)} in this paper.
We also introduce a suitable training procedure for {\it HW-LSTM}.

\subsection{Variants of Highway LSTM}
Unlike a regular RNN, an LSTM has two internal states, memory cell $c$ and hidden state $h$.
We explore three variants, {\it HW-LSTM-C}, {\it HW-LSTM-H}, and {\it HW-LSTM-CH}, which differ in whether the highway network is added to $c$ and/or $h$ as below:

\begin{description}
\item[HW-LSTM-C]
{\it HW-LSTM-C} adds a highway network on top of the memory cell as shown in \figref{fig:hlstm_c}.
Different from a regular LSTM described in \secref{sec:lstm}, $c$ and $h$ are calculated as follows:
	   \begin{eqnarray}
 \acute{c}_{t} &=& c_{t-1} \odot f_{t} + i_{t} \odot j_{t} \nonumber \\
 g_{T} &=& \mbox{sigm}(W_{T}\acute{c}_{t}+b_{T}) \nonumber \\
 g_{C} &=& \mbox{sigm}(W_{C}\acute{c}_{t}+b_{C}) \nonumber \\
 \acute{\acute{c}}_{t} &=& \mbox{tanh}(W\acute{c}_{t}+b) \nonumber \\	    
 c_{t} &=& \acute{c}_{t} \odot g_{C} + \acute{\acute{c}}_{t} \odot g_{T} \nonumber \\
  h_{t} &=& \tanh(c_{t}) \odot o_{t} \nonumber
	   \end{eqnarray}
 \item[HW-LSTM-H]
{\it HW-LSTM-H} adds a highway network on top of the hidden state as in \figref{fig:hlstm_h}.
Only the calculation of $h$ is different from a regular LSTM as below:
\begin{eqnarray}
\acute{h}_{t} &=& \tanh(c_{t}) \odot o_{t} \nonumber \\
g_{T} &=& \mbox{sigm}(W_{T}\acute{h}_{t}+b_{T}) \nonumber \\
 g_{C} &=& \mbox{sigm}(W_{C}\acute{h}_{t}+b_{C}) \nonumber \\
 \acute{\acute{h}}_{t} &=& \mbox{tanh}(W\acute{h}_{t}+b) \nonumber \\
h_{t} &=& \acute{h}_{t} \odot g_{C} + \acute{\acute{h}}_{t} \odot g_{T} \nonumber
\end{eqnarray}
\item[HW-LSTM-CH]
{\it HW-LSTM-CH} adds highway networks on top of both the memory cell and the hidden state as shown in \figref{fig:hlstm_ch}.
	   \begin{eqnarray}
 \acute{c}_{t} &=& c_{t-1} \odot f_{t} + i_{t} \odot j_{t} \nonumber \\
 g_{T}^{c} &=& \mbox{sigm}(W_{T}^{c}\acute{c}_{t}+b_{T}^{c}) \nonumber \\
 g_{C}^{c} &=& \mbox{sigm}(W_{C}^{c}\acute{c}_{t}+b_{C}^{c}) \nonumber \\
 \acute{\acute{c}}_{t} &=& \mbox{tanh}(W_{c}\acute{c}_{t}+b_{c}) \nonumber \\	    
 c_{t} &=& \acute{c}_{t} \odot g_{C}^{c} + \acute{\acute{c}}_{t} \odot g_{T}^{c} \nonumber \\
\acute{h}_{t} &=& \tanh(c_{t}) \odot o_{t} \nonumber \\
g_{T}^{h} &=& \mbox{sigm}(W_{T}^{h}\acute{h}_{t}+b_{T}^{h}) \nonumber \\
g_{C}^{h} &=& \mbox{sigm}(W_{C}^{h}\acute{h}_{t}+b_{C}^{h}) \nonumber \\
\acute{\acute{h}}_{t} &=& \mbox{tanh}(W_{h}\acute{h}_{t}+b_{h}) \nonumber \\
h_{t} &=& \acute{h}_{t} \odot g_{C}^{h} + \acute{\acute{h}}_{t} \odot g_{T}^{h} \nonumber
	   \end{eqnarray}
\end{description}

In the above explanation, for simplicity, the number of highway
network layers was set to one.  We define the number of highway layers
as depth.  Same as described in deep transition networks and recurrent
highway networks~\cite{pascanu2013construct,zilly2016recurrent}, we
can increase the depth in {\it HW-LSTM} by stacking highway network
layers inside the LSTM.



In order to reduce the number of parameters in {\it HW-LSTM-C} and
{\it HW-LSTM-H}, we simplified the carry gate to $g_{C} = 1 - g_{T}$
as in the original paper~\cite{srivastava2015highway}. For {\it
  HW-LSTM-CH}, both carry gates are set $g^{c}_{C} = 1 - g^{c}_{T}$
and $g^{h}_{C} = 1 - g^{h}_{T}$.

\subsection{Training Procedure of Highway LSTM}
\label{sec:train-proc-highw}
Due to the additional highway networks, the training of {\it HW-LSTM} LMs is slower than for regular LSTM LMs.
To mitigate this, we can: (1) train the regular LSTM LM, (2) convert it to {\it HW-LSTM} by adding highway networks, and (3) conduct additional training of {\it HW-LSTM} LM that is converted from the regular LSTM LM.
In other words, {\it HW-LSTM} LM is initialized with the trained regular LSTM LM.
To smoothly convert the regular LSTM LM to the {\it HW-LSTM} LM, we set the bias term for the transformation gate to a negative value, say -3, so that the added highway connection is initially biased toward carry behavior, which means that the behavior of the converted {\it HW-LSTM} LM is almost the same as the regular LSTM LM. Next, we conduct the additional training of {\it HW-LSTM} LM.

In LM training, it is common to have a two-stage training procedure where the first stage uses a large generic corpus and the second stage uses a small specific target-domain corpus.
Our proposed procedure fits this type of two-stage training.

\section{Experiments}
\label{sec:experiments}
We compare the three {\it HW-LSTM} variants, {\it HW-LSTM-C}, {\it HW-LSTM-H}, and {\it HW-LSTM-CH}.
In a first set of experiments, we use English broadcast news data for this comparison and report perplexity and speech recognition accuracy.
Then we conduct experiments with English conversational telephone speech data and compare the speech recognition accuracy with the strong LSTM baseline.
For speech recognition experiments, we generated $N$-best lists from lattices produced by the baseline system for each task and rescored them with the baseline LSTM and/or the {\it HW-LSTM} LMs.
The evaluation metric for speech recognition accuracy was Word Error Rate (WER).
LM probabilities were linearly interpolated and the interpolation weights of LMs were estimated using the heldout data.

\subsection{Baseline LSTM}
\label{sec:baseline-lstm}
\begin{figure}[t]
  \begin{center}
   \includegraphics[width=0.7\columnwidth]{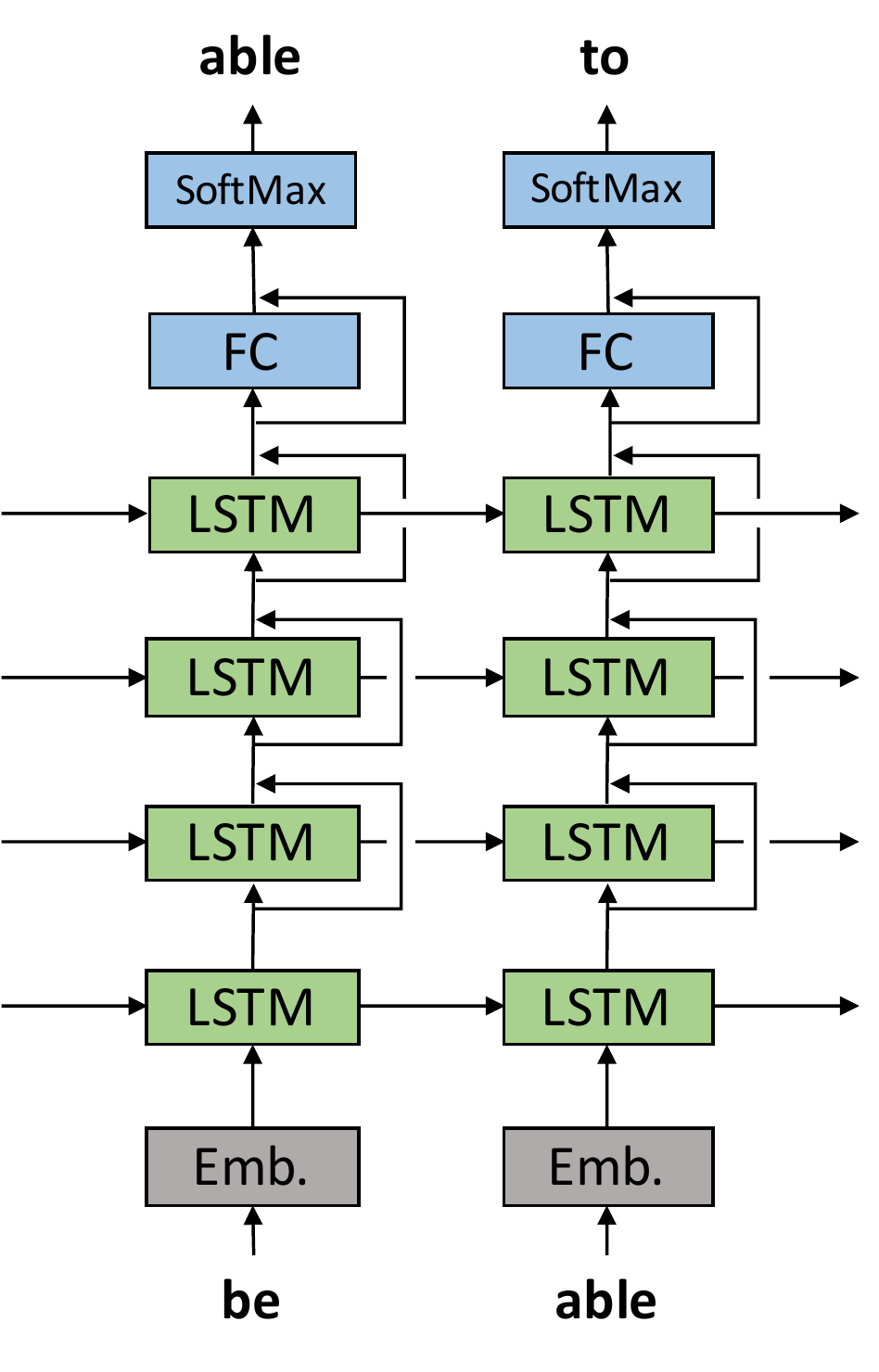}
    \caption{Baseline LSTM LM. ``Emb.'' and ``FC'' indicate word embeddings and fully connected layer.}
  \label{fig:baseline_lstm}
  \end{center}
\end{figure}

Our baseline LSTM LM consists of one word-embeddings layer, four LSTM layers, one fully-connected layer, and one softmax layer, as described in \figref{fig:baseline_lstm}.
The second to fourth LSTM layers and the fully-connected layer allow residual connections~\cite{he2016deep}.
Dropout is applied to the vertical dimension only and not applied to the time dimension~\cite{zaremba2014recurrent}.
This model minimizes the standard cross-entropy objective during training.
The competitiveness of this baseline LSTM LM is detailed in \secref{sec:conv-teleph-speech}

To investigate the advantage of {\it HW-LSTM} LMs, we replaced the LSTM with the {\it HW-LSTM} ({\it HW-LSTM-C}, {\it HW-LSTM-H}, or {\it HW-LSTM-CH}). The rest of the topology is same as the baseline LSTM LM.

\subsection{Network configuration and hyper-parameters}
The baseline LSTM LM uses word embeddings of dimension 256 and 1,024 units in each hidden layer. 
The fully connected layer uses a gated linear unit~\cite{dauphin2016language} and the network is trained with a dropout rate of 0.5.

We used Adam~\cite{kingma2014adam} to control the learning rate in an adaptive way and introduced a layer normalization to achieve stable training for deep LSTM~\cite{ba2016layer}.
In addition to the initial learning rate of 0.001 that is suggested in the original Adam paper~\cite{kingma2014adam}, we tried 0.1 and 0.01.

\subsection{Broadcast news}
Broadcast news evaluation results were reported on the  Defense  Advanced  Research  Projects Agency  (DARPA)  Effective  Affordable  Reusable Speech-to-Text (EARS) RT'04 test set which contains approximately 4 hours of data.  We used two types of acoustic models.
The first model is a discriminatively-trained, speaker-adaptive Gaussian Mixture Model (GMM) acoustic model (AM) trained on 430 hours of broadcast news audio~\cite{chen06:_advan_speec_trans_ears_progr}.
The second model is a Convolutional Neural Net (CNN) acoustic model trained on 2,000 hours of broadcast news, meeting, and dictation data with noise based data augmentation.
The CNN-based AM was first trained with cross-entropy training~\cite{rumelhart1988learning} and then with Hessian-free state-level Minimum Bayes Risk (sMBR) sequence training~\cite{kingsbury2009lattice,kingsbury2012scalable}.  


We trained a conventional word 4-gram model using a total of 350M words from multiple sources~\cite{An.Empirical.Study.of.Smoothing.Techniques.for.Language.Modeling} with a vocabulary size of 84K words.
To compare the baseline LSTM LM and the three types of the proposed {\it HW-LSTM} LMs, we used a 12M-word subset of the original 350M-word corpus, as done in~\cite{arisoy2015bidirectional}.
For reference, we also trained model-M from the same training data with the baseline LSTM~\cite{chen:2009:NAACLHLT091,chen:2009:NAACLHLT092,chenscaling}.
The hyper-parameters were optimized on a heldout data set.

\tabref{tab:ppl_bn} illustrates the perplexity of these models on the heldout set.
The {\it HW-LSTM-H} and {\it HW-LSTM-CH} achieved the best perplexity, whereas the {\it HW-LSTM-C} saw a marginal degradation compared with the baseline LSTM.

\begin{table}[t]
 \begin{center}
  \caption{Perplexity on broadcast news with various LMs.}
  \label{tab:ppl_bn}
  \begin{tabular*}{\columnwidth}{@{\extracolsep{\fill}}lc}
   \thline
   ~&Perplexity \\
   \thline
   $n$-gram & 123 \\
   model-M~\cite{chenscaling} & 121 \\
   Baseline LSTM        & 114 \\
   \hline
   HW-LSTM-C   & 115 \\
   HW-LSTM-H   & 102 \\
   HW-LSTM-CH & 102 \\
   \thline
  \end{tabular*}
 \end{center}
	\end{table}

\begin{table}[t]
 \begin{center}
  \caption{Word Error Rate (WER) on broadcast news after various configurations of LM rescoring.}
  \label{tab:wer_bn}
  \begin{tabular*}{\columnwidth}{@{\extracolsep{\fill}}lcc}
   \thline
   ~& \multicolumn{2}{c}{WER~[\%]} \\
   ~& GMM AM & CNN AM \\
   \thline
   $n$-gram &  13.0 & 10.9 \\
   $n$-gram + Baseline LSTM        & 12.2 & 10.2 \\
   \hline
   $n$-gram & ~ & ~ \\
   ~~+ HW-LSTM-C  & 12.2 & 10.3 \\
   ~~+ HW-LSTM-H  & {\bf 12.0} & 10.1 \\
   ~~+ HW-LSTM-CH & 12.2 & 10.1 \\
   \hline
   $n$-gram + Baseline LSTM & ~ & ~ \\
   ~~+ HW-LSTM-C  & 12.1 & 10.1 \\
   ~~+ HW-LSTM-H  & {\bf 12.0} & {\bf 10.0} \\
   ~~+ HW-LSTM-CH & 12.1 & {\bf 10.0} \\
   \thline
   \end{tabular*}
  \begin{flushright}
   $\ast$~ {\bf Bold} numbers is the best WER for each AM.
  \end{flushright}
 \end{center}
	\end{table}


\tabref{tab:wer_bn} illustrates the WER on EARS RT'04 obtained by rescoring $N$-best lists produced by the two acoustic models.
For reference, the first section in \tabref{tab:wer_bn} compares the WER with $n$-gram LM and the WER after rescoring with the baseline LSTM over the lattices generated with the $n$-gram LM.
As can be seen, rescoring with the baseline LSTM LM significantly reduces WER for both the GMM AM and the CNN AM.
The second section describes the rescoring results by three types of {\it HW-LSTM} LMs over the lattice generated by $n$-gram.
The third section describes the rescoring results by the same {\it HW-LSTM} LMs, but after rescoring by the baseline LSTM LM.
Comparing the first and the second section, {\it HW-LSTM-H} showed better WER than the baseline LSTM both for GMM AM and CNN AM.
When looking at the second and third sections, using {\it HW-LSTM-H} resulted in the best WER both for GMM AM and CNN AM.
While {\it HW-LSTM-H} and {\it HW-LSTM-CH} had similar perplexity, {\it HW-LSTM-H} showed slightly better WER than {\it HW-LSTM-CH}.
{\it HW-LSTM-H} has a smaller number of parameters than {\it HW-LSTM-CH} and thus is less prone to overfitting.

In the experiments with English conversational telephone speech recognition described in the next section, we will use the pipeline of using both the baseline LSTM and {\it HW-LSTM-H} that achieved the best WER in these broadcast news experiments.

\begin{table}[t]
 \begin{center}
  \caption{Discription for test sets for English conversational telephone speech.}
  \label{tab:testset}
   \begin{tabular*}{\columnwidth}{@{\extracolsep{\fill}}lccc}
    \thline
    ~ & Duration & \# speakers & \# words \\
    \thline
SWB & 2.1h & 40 & 21.4K \\
CH  & 1.6h & 40 & 21.6K \\
RT'02       & 6.4h &120 & 64.0K \\
RT'03       & 7.2h &144 & 76.0K \\
RT'04       & 3.4h &72  & 36.7K \\
DEV'04f     & 3.2h &72  & 37.8K \\
       \thline
\end{tabular*}
 \end{center}
\end{table}
\begin{table*}
 \begin{center}
  \caption{Word Error Rate (WER) on conversational telephone speech with various LM configurations.}
  \label{tab:wer_swb}
  \begin{tabular*}{\textwidth}{@{\extracolsep{\fill}}lcccccc}
   \thline
   ~&\multicolumn{6}{c}{WER~[\%]} \\
   ~& SWB & CH & RT'02 & RT'03 & RT'04 & DEV'04f \\
   \thline
   $n$-gram + model-M~\cite{george17:_englis_conver_telep_speec_recog_human_machin} & 6.1 & 11.2 & 9.4 & 9.4 & 9.0 & 8.8 \\
   $n$-gram + model-M + 4 LSTM + CNN~\cite{george17:_englis_conver_telep_speec_recog_human_machin,kurata17:_empir_explor_novel_archit_objec_languag_model} & 5.5 & 10.3 & 8.3 & 8.3 & 8.0 & 8.0 \\
   \hline
   $n$-gram + model-M + Baseline LSTM & 5.4 & 10.1 & 8.4 & 8.3 & 8.0 & 8.1 \\
   \hline
   $n$-gram + model-M + Baseline LSTM & ~ & ~ & ~ & ~ & ~ & ~ \\
   ~~+ HW-LSTM-H (d=1) & 5.3 & 10.1 & 8.3 & 8.3 & 8.0 & 8.1 \\
   ~~~~+ HW-LSTM-H (d=2) & 5.3 & 10.1 & 8.2 & 8.2 & 7.9 & 7.9 \\
   ~~~~~~+ HW-LSTM-H (d=3) & 5.3 & 10.0 & 8.1 & 8.2 & 7.8 & 7.9 \\
   ~~~~~~~~+ HW-LSTM-H (d=4) & 5.3 & 10.0 & 8.1 & 8.2 & 7.8 & 7.9 \\
   ~~~~~~~~~~+ HW-LSTM-H (d=5) & 5.3 & 10.0 & 8.1 & 8.2 & 7.8 & 7.9 \\
   ~~~~~~~~~~~~+ HW-LSTM-H (d=6) & 5.3 &  9.9 & 8.2 & 8.1 & 7.8 & 7.9 \\
   ~~~~~~~~~~~~~~+ HW-LSTM-H (d=7) & 5.2 & 10.0 & 8.1 & 8.1 & 7.8 & 7.9 \\
   \hline
   ~~~+ Unsupervised LM Adaptation & 5.1 & 9.9 & 8.2 & 8.1 & 7.7 & 7.7 \\
   \thline
  \end{tabular*}
 \end{center}
	\end{table*}

\subsection{Conversational telephone speech}
\label{sec:conv-teleph-speech}
To confirm the advantage of {\it HW-LSTM}, we conducted experiments with a wide range of conversational telephone speech recognition test sets, including SWB and CH subsets of the NIST Hub5 2000 evaluation data set and also the RT'02, RT'03, RT'04, and DEV'04f test sets of DARPA-sponsored Rich Transcription evaluation. Statistics of these six data sets are described in \tabref{tab:testset}.

The acoustic model uses an LSTM and a ResNet~\cite{he2016deep} whose posterior probabilities are combined during decoding~\cite{george17:_englis_conver_telep_speec_recog_human_machin}.

The baseline LSTM LM and the {\it HW-LSTM-H} LMs were built with a vocabulary of 85K words.
In the first pass, the LMs were trained with the corpus of 560M words consisting of publicly available text data from LDC, including Switchboard, Fisher, Gigaword, and Broadcast News and Conversations.
In a second pass, this model was refined further with just the acoustic transcripts (approximately, 24M words) corresponding to the 1,975 hour audio data used to train the acoustic models~\cite{xiong2016achieving}.
The hyper-parameter was optimized on a heldout data set.

We tried {\it HW-LSTM-H} LMs with different depths from 1 to 7.
Training of {\it HW-LSTM-H} LMs is slower than that of the baseline LSTM LMs because of their additional highway connections.
Thus, we used the training procedure introduced in \secref{sec:train-proc-highw}.
In the first pass with larger corpora, we trained the baseline LSTM LM.
Then we added the highway connections to the trained baseline LSTM LM to compose the {\it HW-LSTM-H}.
To smoothly convert the baseline LSTM LM to the {\it HW-LSTM-H} LM, we set the bias term for the transformation gate to a negative value, -3, so that the added highway connection was initially biased toward carry behavior which means that the behavior of the composed {\it HW-LSTM-H} LM was almost the same with that of the trained baseline LSTM LM.
Then we conducted the second pass to further train {\it HW-LSTM-H} LM by only using the acoustic transcripts.

Again, the interpolation weights when interpolating the baseline LSTM LM and the {\it HW-LSTM-H} LMs with different depths were optimized on a heldout data set.

The WERs on the six test sets are tabulated in \tabref{tab:wer_swb}.
The first section shows the reference from previous papers~\cite{george17:_englis_conver_telep_speec_recog_human_machin,kurata17:_empir_explor_novel_archit_objec_languag_model}.
In the case of ``$n$-gram + model-M'', lattices were generated using an $n$-gram LM and rescored with model-M~\cite{chen:2009:NAACLHLT091,chen:2009:NAACLHLT092,Saon+2016}.
``$n$-gram + model-M + 4 LSTM + CNN'' line indicates the previously reported results~\cite{george17:_englis_conver_telep_speec_recog_human_machin} that achieved the state-of-the-art WER in the SWB and CH test sets.

The second section of ``$n$-gram + model-M + Baseline LSTM'' is the baseline in this paper.
$n$-gram and model-M are identical with the ones in the first section, however, our baseline LSTM explained in \secref{sec:baseline-lstm} has a different architecture from the LSTM in the first section.
Note that the WERs obtained in ``$n$-gram + model-M + Baseline LSTM'' are comparable with the WERs in ``$n$-gram + model-M + 4 LSTM + CNN'', which indicates that our baseline in this paper is sufficiently competitive.

The third section is our main results and we conducted rescoring by {\it HW-LSTM-H} LMs over the lattices prepared in the second section (``$n$-gram + model-M + Baseline LSTM'').
Here, we incrementally applied the deeper {\it HW-LSTM-H} as $d=1$ to $d=7$.
Note that $d$ indicates the number of the highway networks in {\it HW-LSTM-H} and the number of {\it HW-LSTM-H} layers were kept unchanged to four in all cases.
While there are a marginal number of exceptions, adding deeper {\it HW-LSTM-H} gradually reduces WERs in all test sets.
Comparing with our competitive baseline in this paper, after adding {\it HW-LSTM-H (d=7)}, we obtained absolute 0.2\%, 0.1\%, 0.3\%, 0.2\%, 0.2\%, 0.2\% WER reductions respectively for SWB, CH, RT'02, RT'03, RT'04, and DEV'04f test sets.

Finally, in the fourth section, we conducted an unsupervised LM adaptation after rescoring by {\it HW-LSTM-H} LMs.
We started with re-estimation of the interpolation weights using the rescored results for each test set as a heldout set and conducted rescoring again.
Then, we adapted the model-M LM trained only from acoustic transcripts with using the rescored results for each test set obtained in the previous step~\cite{chen:2009:NAACLHLT092}.
We rescored the $N$-best lists with the adapted model-M and obtained the final results.
After all of unsupervised LM adaptation steps, we reached 5.1\% and 9.9\% WER for SWB and CH subsets of the Hub5 2000 evaluation.


\section{Conclusion}
\label{sec:conclusion}
In this paper, we proposed a language modeling with using {\it HW-LSTM} and confirmed its advantage by conducting a range of speech recognition experiments.
While it may appear that the gains are marginal, it is quite significant at these low WER scenarios obtained with strong LMs and is typical when working with very strong baselines (e.g. 0.2\% improvement for the SWB subset was statistically significant at $p=0.03$\footnote{Statistical significance was measured by the Matched Pair Sentence Segment test~\cite{gillick1989some} in {\tt sc\_stats}.}.).

It is noteworthy that 5.1\% and 9.9\% WER for SWB and CH subsets of the Hub5 2000 evaluation reach the best reported results~\cite{xiong17:_micros_conver_speec_recog_system,george17:_englis_conver_telep_speec_recog_human_machin,kurata17:_empir_explor_novel_archit_objec_languag_model} to date with being achieved by the same system architecture for both tasks.
While a wide range of discussion on the human performance of speech recognition is ongoing~\cite{xiong2016achieving,xiong17:_micros_conver_speec_recog_system,george17:_englis_conver_telep_speec_recog_human_machin,xiong2017microsoft,stolcke2017comparing}, the achieved WER of 5.1\% for the SWB subset is on par with one of the recently estimated human performance for this subset~\cite{george17:_englis_conver_telep_speec_recog_human_machin}.

We conclude:

\begin{itemize}
 \item Among the three variants of {\it HW-LSTM}, {\it HW-LSTM-H} that adds highway network to a hidden state inside LSTM is superior to {\it HW-LSTM-C} and {\it HW-LSTM-CH}\footnote{Comparing with a different combination of highway connection and LSTM that uses a different gating mechanism~\cite{Irie+2016} is worth trying.}.
 \item {\it HW-LSTM-H} LM reduces WER over the strong baseline LSTM LM in English broadcast news and a wide range of conversational telephone speech recognition tasks.
 \item Baseline LSTM LM and {\it HW-LSTM-H} are complementary and a combination of them can result in reduction in WER.
\end{itemize}



\section{Acknowledgment}
We would like to thank Michael Picheny, Stanley F. Chen, and Hong-Kwang J. Kuo of IBM T.J. Watson Research Center for their valuable comments and supports.



\bibliographystyle{IEEEbib}
\bibliography{../../BIB/BIB4INTERNATIONAL}

\end{document}